\documentclass[10pt,journal,compsoc]{IEEEtran}
\ifCLASSOPTIONcompsoc
  \usepackage[nocompress]{cite}
\else
  \usepackage{cite}
\fi
\ifCLASSINFOpdf
\else
\fi

\usepackage{tikz}
\usepackage{comment}
\usepackage{amssymb}
\usepackage{amsmath}
\usepackage{booktabs}
\usepackage{multicol}
\usepackage{color}
\usepackage{xspace}
\usepackage{xcolor}
\usepackage{subcaption}
\usepackage{graphicx}

\usepackage{mathrsfs}
\usepackage{colortbl}
\usepackage{tabularx}
\usepackage{color}
\usepackage{bm}
\usepackage{multirow}
\usepackage{hyperref}       
\usepackage{url}            
\usepackage{booktabs}       
\usepackage{amsfonts}       
\usepackage{nicefrac}       
\usepackage{microtype}      

\usepackage{xcolor}         

\usepackage{subcaption}
\usepackage{makecell}
\usepackage{pdfpages}
\usepackage{enumitem}
\usepackage{gensymb}
\usepackage{wrapfig, lipsum}
\usepackage{ragged2e}
\usepackage{pifont}
\newcommand{\cmark}{\ding{51}}%
\newcommand{\xmark}{\ding{55}}%
\usepackage{xspace}

\newcommand{\tbf}[1]{\textbf{#1}}
\newcommand{\udl}[1]{\underline{#1}}

\newlength\savewidth

\newcommand{\etal}{\textit{et al}.}

\def\ours{\texttt{{DualCoOp++}}\xspace}
\def\ourconf{\texttt{{DualCoOp}}\xspace}

\begin{document}
\title{DualCoOp++: Fast and Effective Adaptation to Multi-Label Recognition with Limited Annotations}

\author{Ping~Hu,
    ~~Ximeng~Sun,
    ~~Stan Sclaroff,
    ~and~Kate~Saenko
\IEEEcompsocitemizethanks{\IEEEcompsocthanksitem P. Hu is with the School of Computer Science and Engineering, UESTC, Chengdu, Sichuan, 611731 and the Department
of Computer Science, Boston University, Boston, MA 02215.\protect\\
E-mail: pinghu@bu.edu}
\IEEEcompsocitemizethanks{\IEEEcompsocthanksitem X-M. Sun, and S. Sclaroff are with the Department
of Computer Science, Boston University, Boston, MA 02215.\protect\\
E-mail: pinghu@bu.edu}
\IEEEcompsocitemizethanks{\IEEEcompsocthanksitem K. Saenko are with Boston University, Boston, MA 02215 and Meta, Menlo Park, CA  94025.\protect}
}

\markboth{Transactions on Pattern Analysis and Machine Intelligence}%
{Shell \MakeLowercase{\textit{et al.}}: Bare Demo of IEEEtran.cls for Computer Society Journals}

\IEEEtitleabstractindextext{%
\begin{abstract}
Multi-label image recognition in the low-label regime is a task of great challenge and practical significance. Previous works have focused on learning the alignment between textual and visual spaces to compensate for limited image labels, yet may suffer from reduced accuracy due to the scarcity of high-quality multi-label annotations. In this research, we leverage the powerful alignment between textual and visual features pretrained with millions of auxiliary image-text pairs.  We introduce an efficient and effective framework called \textit{Evidence-guided Dual Context Optimization} (\ours), which serves as a unified approach for addressing partial-label and zero-shot multi-label recognition.
In \ours we separately encode evidential, positive, and negative contexts for target classes as parametric components of the linguistic input (i.e., prompts). The evidential context aims to discover all the related visual content for the target class, and serves as guidance to aggregate positive and negative contexts from the spatial domain of the image, enabling better distinguishment between similar categories. Additionally, we introduce a Winner-Take-All module that promotes inter-class interaction during training, while avoiding the need for extra parameters and costs.
As \ours imposes minimal additional learnable overhead on the pretrained vision-language framework, it enables rapid adaptation to multi-label recognition tasks with limited annotations and even unseen classes. Experiments on standard multi-label recognition benchmarks across two challenging low-label settings demonstrate the superior performance of our approach compared to state-of-the-art methods.
\end{abstract}

}

\maketitle
\IEEEdisplaynontitleabstractindextext
\IEEEpeerreviewmaketitle

\IEEEraisesectionheading{\section{Introduction}\label{sec:introduction}}
\IEEEPARstart{I}mage recognition has become a very popular and successful research area in recent years, due to the development of large-scale datasets~\cite{deng2009imagenet,kuznetsova2020open} and advanced model architectures~\cite{dosovitskiy2020image,he2016deep,liu2021swin,simonyan2014very}. 
However, the majority of image recognition approaches have focused on single-label prediction, which ignores the intrinsic multi-label nature of images.
Unlike single-label recognition~\cite{dosovitskiy2020image,he2016deep,liu2021swin,simonyan2014very}, multi-label image recognition aims to recognize all semantic labels present in an image~\cite{chen2019multi,chua2009nus,liu2017semantic,liu2018multi,sarafianos2018deep,yazici2020orderless,wang2020multi}, providing a more comprehensive understanding and benefiting applications like image retrieval, video analysis, and recommendation systems.

Multi-label recognition (MLR) typically deals with images of complex scenes and diverse objects. Collecting multi-label annotations becomes difficult to scale up, for two reasons: (i) annotating images with the full semantic label set is laborious and (ii) samples of particular categories can be hard to find. 
The first challenge can be addressed by multi-label recognition with \textit{partial labels}, where merely some of the categories are annotated for each training image. 
Recent works 
proposed solutions to partial-label MLR based on semi-supervised learning~\cite{joulin2016learning,mahajan2018exploring}, normalized training objectives~\cite{durand2019learning}, or label correlations~\cite{chen2021structured,huynh2020interactive,pu2022semantic}.
The second setting involves \textit{zero-shot} MLR, where novel unseen categories are recognized by transferring knowledge from seen categories, with solutions such as principal image features~\cite{ben2021semantic,zhang2016fast}, knowledge graphs~\cite{lee2018multi}, and attention mechanisms~\cite{huynh2020shared,narayan2021discriminative}.
Despite significant progress in the two settings, existing approaches are not designed to handle both at once. We propose to unify these settings as \textit{limited-annotation} MLR and design a solution that can handle practical scenarios with either partial or missing labels.

\begin{figure}[t]
\includegraphics[width=\linewidth]{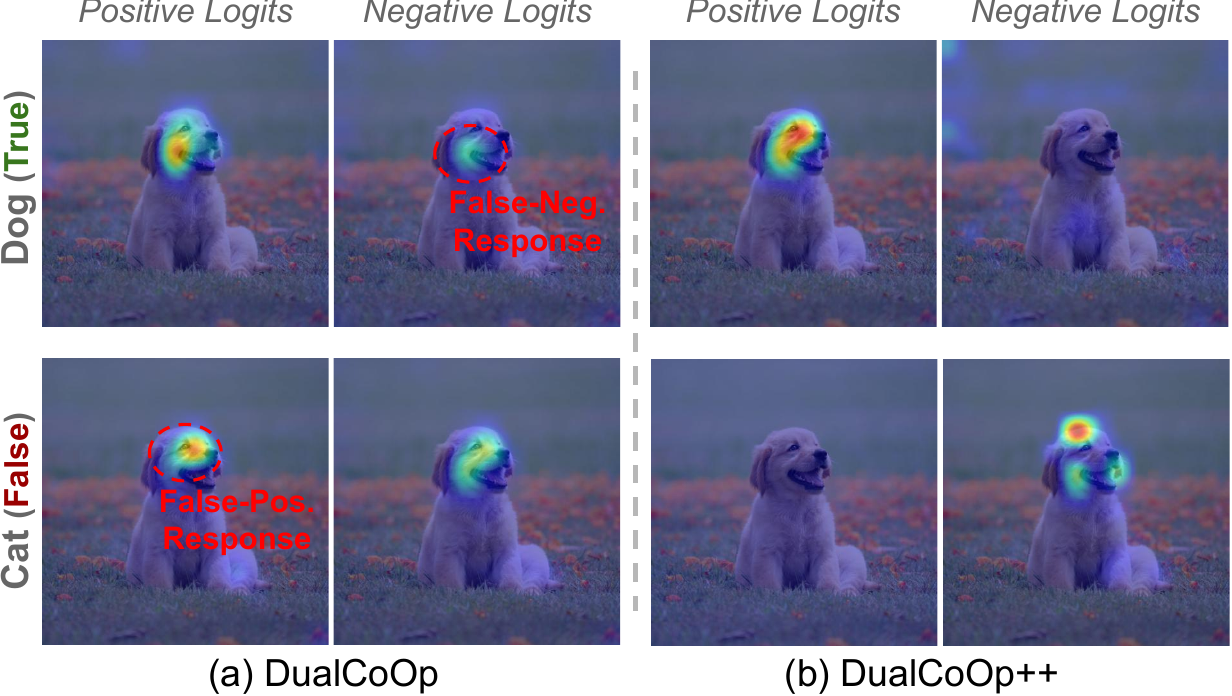}
\caption{\small 
    \textbf{Visualization of positive and negative logit maps for the true label ``Dog'' and the false label ``Cat''.} In contrast to \ourconf showing high false-negative and false-positive activations, \ours presents better abilities to suppress incorrect predictions.
    }
   
\label{fig:teaser}
\end{figure}

\begin{figure*}[t]
\begin{center}
     \includegraphics[width=\linewidth]{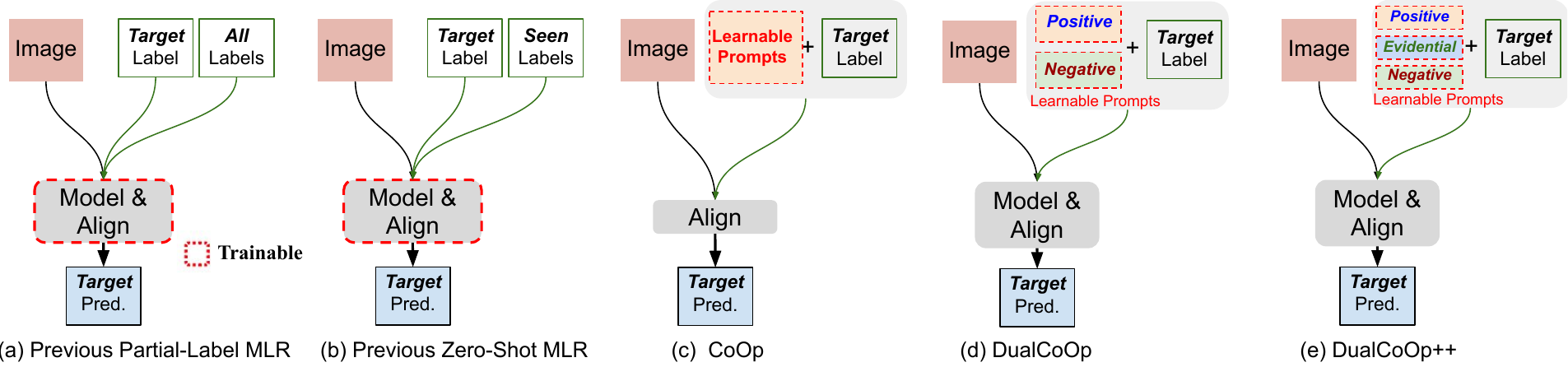}
\end{center} 
   \caption{\small 
   \textbf{A conceptual comparison of previous multi-label recognition (MLR) methods and our approaches}. In Partial-Label MLR (a) and Zero-Shot MLR (b), previous works learn to model and align the visual and textual inputs as well as explore the correlation between the target label with all/seen labels depending on the limited semantic annotations available on the dataset, which leads to sub-optimal performance and complex model architectures. 
   Large-scale pretrained vision-language models like CoOp~\cite{zhou2022conditional} (c) can address MLR by aligning visual and textual inputs but learn for only positive prediction and overlook the fine-grained visual spatial details crucial for multi-label image recognition. In contrast, we propose a unified framework (d) to address the limitations and tackle both limited-annotation tasks ~\cite{sun2022dualcoop}, and further improve the representation capability (e). With the addition of a set of lightweight learnable prompts, the proposed methods effectively recognize different classes within complex visual scenes.  }
   \label{fig:concept}
\end{figure*}

Successful solutions to the above problems transfer knowledge from fully-annotated categories to partially-labeled and novel categories by learning an alignment between images and category names~\cite{chen2021structured,pu2022semantic,zhang2016fast}. 
Recently, vision-language pretraining models are bridging the visual-textual gap via large-scale pretraining, e.g., CLIP~\cite{radford2021learning} is trained with 400 million image-text pairs. In this work, we draw inspiration from the recent success of prompt learning for such models~\cite{huang2022unsupervised,ju2021prompting,luddecke2021mm,radford2019language,zhou2021denseclip,zhou2022conditional,zang2022unified,ma2023understanding,liu2023hierarchical,wasim2023vita}. 
Prompt learning provides a convenient way to transfer pretrained vision-language models to other tasks. 
It designs additional templated or learnable prompt tokens for textual input to “inform” the model about downstream tasks and avoids finetuning the entire model which can be inefficient and data-hungry. 
By doing so, recent works like CoOp~\cite{zhou2022conditional} have demonstrated CLIP's remarkable generalization to various zero-shot image tasks~\cite{huang2022unsupervised,radford2021learning,zhou2022conditional} (Fig.~\ref{fig:concept} (c)).
However, these methods mainly focus on matching each image with a single label, hence they are not able to handle the multi-label setting.

To adapt the knowledge learned in CLIP to multi-label image recognition, we propose the \ourconf in the conference version~\cite{sun2022dualcoop} of this work. 
As shown in Fig.~\ref{fig:concept} (d), \ourconf learns a pair of differentiable prompts to provide positive and negative contexts for the target class. 
Rather than focusing solely on positive predictions~\cite{zhou2022conditional}, the dual prompts naturally give rise to two independent and complementary classifiers for positive and negative predictions. In this way, the method facilitates more balanced learning from different annotation types in multi-label image recognition and also eliminates the need for manually setting thresholds for classification~\cite{ridnik2021asymmetric}.
In contrast to previous models depicted in Fig.~\ref{fig:concept} (a)(b), our proposed method avoids fine-tuning the entire vision-language model. Instead, it only trains on the prompts, which are notably smaller in scale compared to the entire model. As a result, this simple framework attains significantly higher efficiency when adapting to diverse datasets.

In \ourconf~\cite{sun2022dualcoop}, we present the Class-Specific Region Feature Aggregation, wherein the positive correlation is directly normalized to serve as spatial attention for aggregating final positive and negative logits. For samples with true labels (i.e., the image contains the target class), this design effectively promotes true-positive predictions by accentuating positive logits and suppressing negative logits.
When dealing with false labels (i.e., the image lacks the target class), the objective shifts to optimizing true-negative predictions by minimizing positive logits and maximizing negative logits. However, minimizing positive logits in this context may hinder the positive correlation map's responsiveness, which is crucial for identifying confusing regions. This, in turn, leads to distracted aggregation weights that complicate the learning of negative logits, potentially resulting in a loss of accuracy. To address this limitation, this extended version of the work introduces \ours, incorporating an Evidence-Guided Region Feature Aggregation module to disentangle the correlation map from positive logits.
As shown in Fig.~\ref{fig:concept}(d), besides the positive and negative contexts, we further introduce the evidential context to guide the spatial aggregation of positive and negative contexts.
Unlike positive logits and negative logits that directly indicate the existence and non-existence of object classes, evidential logits aim to extract all the related visual contents showing similar representations. As a result, optimizing the positive branch will not affect the learning of the negative branch, and the model can better represent and distinguish between target classes and similar classes as visualized in Fig.~\ref{fig:teaser}. 
Moreover, since \ourconf is optimized for each class individually, the lack of interaction among classes may also hamper the performance. In particular, an image region can positively respond to multiple similar classes, hence resulting in false-positive predictions. To address this challenge, we further propose a Winner-Take-All (WTA) module that regularizes each spatial location to only positively respond to at most one category, thus further enhancing the model's ability to distinguish between similar categories. 
Since WTA is a non-parametric module, the proposed framework keeps high efficiency without introducing extra computational overhead. With these design choices, we achieve a unified framework for addressing the general challenges of multi-label recognition with limited annotations. 

Our contributions can be summarised as follows:
\begin{itemize} [leftmargin=*]
    \item We propose \ours to efficiently and effectively adapt a powerful vision-language model to solve multi-label recognition tasks using limited annotations.
     \item We propose the Evidence-Guided Region Feature Aggregation module to improve the spatial aggregation of contextual information learned from limited annotations.
    \item We propose the Winner-Take-All (WTA) module to promote the inter-class interaction in MLR, leading to better distinguishing similar categories.
    \item We conduct extensive experiments and analysis on multiple benchmark datasets, and demonstrate that \ours achieves the state-of-the-art performance of both partial-label MLR  and zero-shot MLR. Notably, without introducing extra computational overhead, \ours consistently improves our previous \ourconf by more than 2$\%$ for both tasks on benchmarks like MS-COCO and NUS-WIDE.
\end{itemize}

\section{Related Works}
\textbf{Multi-Label Recognition with Limited Annotations.}
Multi-label image recognition has drawn increasing attention in recent years. 
One straightforward solution to this problem is to individually learn a binary classifier for each category ~\cite{liu2015optimality,misra2016seeing,tsoumakas2007multi}, which however does
not consider correlations among labels. 
Hence, recent works have focused on incorporating semantic dependencies among labels via graph neural networks~\cite{chen2019multi,chua2009nus,wang2020multi} or RNN/LSTM~\cite{liu2017semantic,wang2016cnn,wang2017multi,yazici2020orderless}. 
Some work also considers the spatial distribution of labels in the image, and exploits object proposals~\cite{li2016human,liu2018multi,wang2016beyond} or attention mechanism~\cite{sarafianos2018deep,wang2017multi,zhu2017learning} as a regularization to rectify the prediction.
However, despite achieving significant progress, these methods require a large-scale and complete annotated dataset to train models~\cite{krishna2017visual,lin2014microsoft}. 
This limits their application to more practical scenarios where the data is partially annotated for training~\cite{bucak2011multi,chen2013fast,mahajan2018exploring,sun2017revisiting,sun2010multi,xie2018partial} and unseen (zero-shot) categories may appear during testing~\cite{chen2020knowledge,gupta2021generative,lee2018multi,mensink2014costa,zhang2016fast}. 

With partially labeled data, where merely some labels of each sample are known, Mahajan~\etal~\cite{mahajan2018exploring} and Joulin~\etal~\cite{joulin2016learning} attempt to use web supervision to automatically generate the pseudo labels, which unfortunately leads to poor performance as the web supervision is noisy and incomplete~\cite{zhang2021understanding}. To avoid external noise, Durand~\etal~\cite{durand2019learning} exploit the proportion of annotated samples for different labels and propose a normalized BCE loss to train models based on the given partial labels. 
More recent works explicitly transfer information from known
labels to complement unknown labels by utilizing  category-specific feature blending~\cite{pu2022semantic} or label co-occurrences~\cite{chen2021structured} at both instance-level and prototype-level.

Unlike partial annotation of the same label set for training and testing, zero-shot multi-label image recognition needs to handle novel categories during testing, hence inspiring a different route based on a joint visual-label embedding space~\cite{chen2020knowledge,huynh2020shared,mensink2014costa,narayan2021discriminative,zhang2016fast}. 
Zhang~\etal~\cite{zhang2016fast}
propose to find a  principal direction that ranks related labels first in the joint embedding space optimized via a tailored zero-shot ranking loss. Cohen~\etal~\cite{ben2021semantic} further improve the idea by learning multiple principal vectors to support the semantic diversity. 
Huynh~\etal~\cite{huynh2020shared} consider the spatial regularization and propose a shared multi-attention model and obviate the need for explicit region proposals~\cite{rahman2019deep0tag}. 
Narayan~\etal~\cite{narayan2021discriminative} propose to enhance the region-based features so as to minimize inter-class feature entanglement.

Though significant progress has been made in each of the directions, existing methods still require a lot of MLR data and complex architectures/losses. Our approach reduces the need for hard-to-get MLR data by pretraining on unsupervised text-image pairs. While it may seem unfair to compare existing MLR methods with ones based on such pretraining, we point out that the pretraining data is unsupervised and thus easier to obtain. We also provide experiments that compare \ours to baselines using the same
pretraining.
Importantly, previous methods are designed for only one task, and hence have limitations in practical applications. In contrast, our proposed framework can be easily adapted to small data and can address both partial and zero-shot tasks at the same time.   

\vspace{1mm}
\textbf{Prompt Learning for Vision-Language Models.}
Vision-Language Models~\cite{jia2021scaling,radford2021learning} based on contrastive learning have demonstrated an impressive ability to learn generic visual representations. As a milestone, CLIP~\cite{radford2021learning} is trained with 400 million curated image-text pairs, and shows remarkable transfer capability for over 30 classification datasets.
With such powerful vision-language models, several follow-ups~\cite{gao2021clip,wortsman2021robust,yao2021cpt,zhang2021tip,derakhshani2022variational,liu2023patch,miyai2023locoop} have been proposed to explore the training strategies for training downstream classification tasks.
Instead of fine-tuning the entire model~\cite{dong2019unified,he2016deep}, which may damage the learned representation space, recent approaches adopt  the prompt-based paradigm that formalizes NLP tasks as masked language modeling (prompt templates)~\cite{lester2021power,li2021prefix,shin2020autoprompt,zhang2023vision,yu2023visual}.
Zhou~\etal~\cite{zhou2021learning} propose to tune prompts for downstream classification tasks, and further introduce input-conditional prompts for better generalization ability~\cite{zhou2022conditional}. Lu~\etal~\cite{lu2022prompt} learn the distribution of diverse prompts to handle the varying visual representations. Huang~\etal~\cite{huang2022unsupervised} generate pseudo labels for images to learn prompts in an unsupervised way.
Though achieving promising improvements for downstream tasks, these methods address the multi-class zero-shot image recognition, assuming each image has one label, hence lacking the ability to handle the multi-label setting, especially under the low-label regime.  
Toward this direction, Ding~\etal~\cite{ding2023exploring} introduce a semantic correspondence prompt network to enhance the semantic context with label-to-label semantic priors, while Guo~\etal~\cite{guo2023texts} exploit rich text description data to learn stronger prompts. In contrast, without relying on extra networks or data, we propose a unified method that transfers vision-language models to address limited-annotation multi-label image recognition with better performance.

\section{Method}
\vspace{1mm}
\textbf{Problem Definition.}
We formally define multi-label recognition with limited annotations as follows: Consider $M$ as the set of categories that describe objects or attributes in images. Given a training image $I$, the existence of a category $m\in M$ can be positive, negative or unknown, corresponding to the label $y_m = 1, -1$ or $0$ respectively. During inference, we predict each label of interest for an input image. 

Many existing MLR problems fit into this broad definition. In this paper, we consider the settings with partial or missing labels: (1) \textbf{Partial-label MLR}~\cite{chen2021structured,durand2019learning,pu2022semantic}, in which only a subset of labels are known ($+1$ or $-1$) for each training image and we are interested in predicting all existing labels during inference. (2) \textbf{Zero-shot MLR}~\cite{ben2021semantic, huynh2020shared, rahman2018deep}, in which each label is either known (seen) or unknown (unseen) for \textit{all} images during training and we are interested in predicting either all labels or only unknown (unseen) labels during inference. 
In this paper, we propose a unified setting that includes both scenarios, which we call  \textit{limited-annotation MLR}.

\begin{figure*}
\begin{center}
     \includegraphics[width=0.95\linewidth]{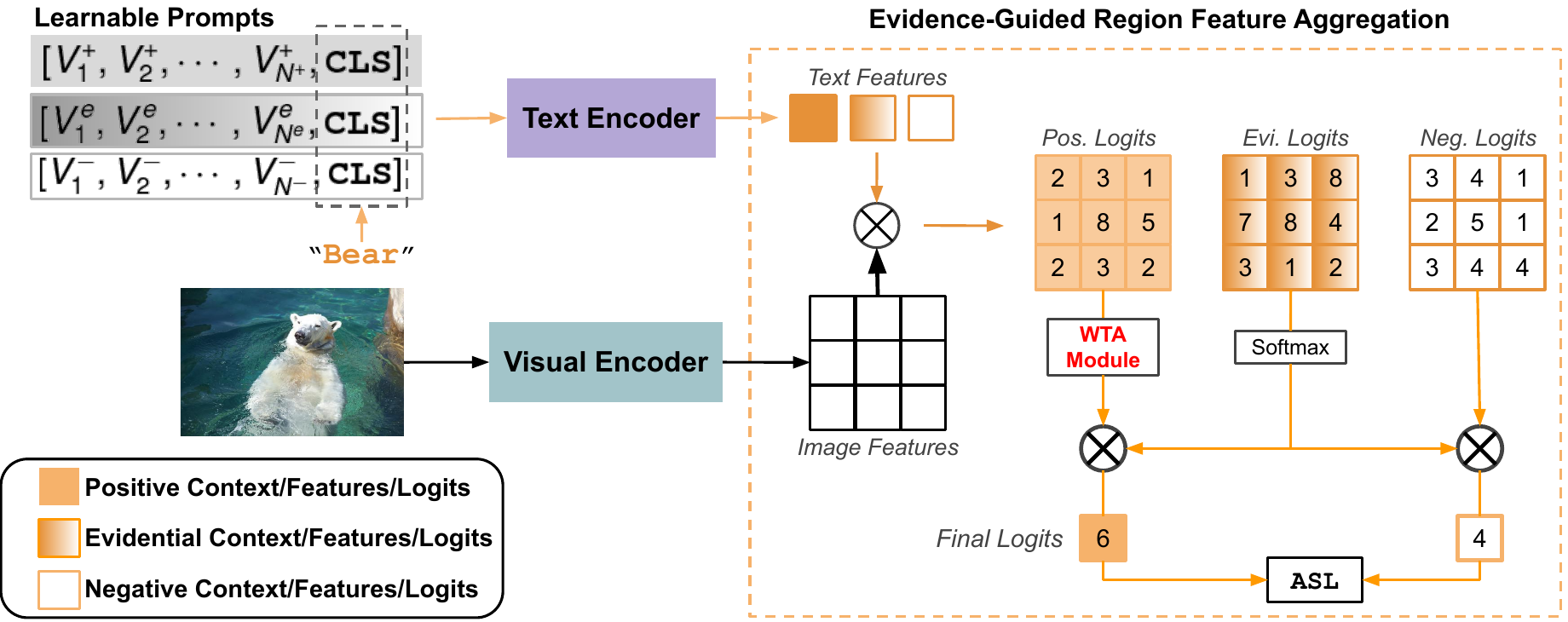}
\end{center} 
   \caption{\small 
   \textbf{Illustration of our proposed approach.} \ours learns a triplet of evidential,  positive, and negative prompts to quickly adapt powerful pretrained vision-text encoders to the MLR task. For each class, three prompts generate three contrastive (evidential, positive, and negative) textual embeddings as the input to the text encoder. Furthermore, we propose \textit{Evidence-Guided Region Feature Aggregation} which projects each region's feature to the textual space first and then aggregates the spatial logits by the magnitude of class-specific evidential responses. A winner-take-all (WTA) module is also utilized to promote cross-class interaction.
   During training, we apply the ASL loss~\cite{ridnik2021asymmetric} to optimize learnable prompts while keeping other network components frozen. During inference, we compare the final positive and negative logits to make a prediction for each class.
   }\label{fig:model_overview}
   \vspace{-10pt}
\end{figure*}

\vspace{1mm}
\noindent\textbf{Approach Overview.}
To compensate for insufficient or missing image labels, it is important to learn how the meanings of category names are related to each other, so that we can transfer knowledge between related categories. This is usually done by learning an alignment between the visual and textual spaces. However, our dataset is too limited to learn a broad and generalizable mapping. 
Instead, we propose to leverage the strong alignment of visual and textual feature spaces learned by large-scale vision-language pretraining (CLIP~\cite{radford2021learning}) with a light-weight learnable overhead that quickly adapts to the MLR task with limited semantic annotations.  Figure~\ref{fig:model_overview} provides an overview of our proposed approach. \ours learns a triplet of ``prompt'' contexts in the form of three learnable sequences of word vectors, to provide evidential, positive, and negative contexts of a given category name. This generates evidential, positive, and negative textual features that are fed into the pretrained text encoder. 
To better distinguish and recognize target objects, which can be located at different locations in the image and similar to other categories, we introduce an evidence-guided spatial aggregation step. We first compute the similarity score of the projected visual feature maps with the three context encodings to obtain three prediction logits over each region. For each class,  we perform spatial aggregation of all positive/negative logits, in which the weight for each logit is determined by its relative magnitude of the evidential logits, and we call this \textit{Evidence-Guided Region Feature Aggregation}. By doing so, the proposed framework can be more flexible to represent and distinguish the target class and similar classes (e.g. Regions of ``Dog'' can show high similarity to label ``Cat'' in the evidential logit map, while avoiding response in the positive logit map). We further apply a non-parametric Winner-Take-All module to promote inter-class interaction and optimize the learnable prompts via the ASL loss~\cite{ridnik2021asymmetric} while keeping all other network components frozen. During inference, we directly compare the final positive and negative logits to make a prediction for the label of a category $y$.

\noindent\textbf{Triple Learnable Prompts.} Instead of learning a single~\cite{zhou2021learning} or dual~\cite{sun2022dualcoop} prompt(s) for a class, we propose Evidence-Guided Context Optimization (\ours) which learns three contrastive prompts' contexts for each class. The learnable part in triple prompts carries evidential, positive, and negative contextual surroundings individually and can be optimized end-to-end from data via binary classification loss. Specifically, we define the triplet of prompts given to the text encoder as follows:
\begin{align}
    {P}^{e}~&= \big[V_1^{e},~V_2^{e}, \cdots ,~V_{N^e}^{e}, \texttt{CLS}\big]\\
    {P}^{+} &= \big[V_1^{+},V_2^{+}, \cdots ,V_{N^+}^{+}, \texttt{CLS}\big]\\
    {P}^{-} &= \big[V_1^{-},V_2^{-}, \cdots ,V_{N^-}^{-}, \texttt{CLS}\big]
\end{align}
where each $V$ is a learnable word embedding vector (\textit{e.g.} with dimension 512 in CLIP~\cite{radford2021learning}) and $\texttt{CLS}$ is the given category name. $N^{e}$, $N^{+}$, and $N^{-}$ are the numbers of word tokens learned in the evidential, positive, and negative prompts respectively. For simplicity, we utilize the same size of prompts in our experiments. We learn a triplet of prompts for each class (i.e. class-specific prompt triplet) when solving MLR with partial labels, and learn a triplet of prompts shared for all classes in zero-shot MLR. With a triplet of prompts, we compute the binary classification output $p$ by comparing the positive and negative contexts as:
\begin{align}
    \label{eq:prediction}
    p = \frac{e^{\delta(E_v^I,E_t^e,E_t^+)/\tau}}{e^{\delta(E_v^I,E_t^e,E_t^+)/\tau}+e^{\delta(E_v^I,E_t^e,E_t^-)/\tau}}
\end{align}
where $p$ is the predicted probability for a given (image, label) pair as a positive example, and $\tau$ is a temperature parameter. $E_v^I$ is the visual encoding feature maps. $E_t^e$, $E_t^+$, $E_t^-$ are the textual encodings for evidential, positive, and negative prompts, respectively. $\delta(\cdot,\cdot, \cdot)$ is our proposed evidence-guided spatial aggregation function to adaptively reduce the spatial dimension of visual features for each class, which will be discussed next.

\vspace{1mm}

\noindent\textbf{Evidence-Guided Region Feature Aggregation.} In multi-label image recognition, it is common for multiple objects to appear in different regions of the image. 
Pooling to produce a single image-level feature vector for all classes gives sub-optimal performance since spatial information is reduced and different objects are mixed.
In this work, we reformulate the last multi-headed attention pooling layer of the visual encoder in CLIP~\cite{radford2021learning} and apply evidence-guided class-specific pooling to adaptively aggregate region features in the multi-label setting. The original attention pooling layer in CLIP pools the visual feature map first, and then projects the global feature vector into text space as follows: 
\begin{align}
    \text{AttnPool}(x)  &= \text{Proj}_{v \rightarrow t} (\sum_i \text{softmax}(\frac{q(\bar x) k(x_i)^T}{C}) \cdot v(x_i) ) \nonumber \\
                        &= \sum_i \text{softmax}(\frac{q(\bar x) k(x_i)^T}{C}) \cdot \text{Proj}_{v \rightarrow t} (v(x_i)) \nonumber\\
                        &= \text{Pool}(\text{Proj}_{v \rightarrow t} (v(x_i)) )
\end{align}
where $q$, $v$ and $k$ are independent linear embedding layers and $x = E_v^I$ is the output feature map of the visual encoder. By removing the pooling operation, we can project the visual feature $F_v^i$  of each region $i$ to the textual space~\cite{zhou2021denseclip}:
\begin{align}
    F_v^i = \text{Proj}_{v \rightarrow t} (v(E_i^I))
\end{align}

For each region $i$ and a target class, we compute the logits with cosine similarity between $F_v^i$ and the class's evidential,  positive, and negative contexts,
\begin{align}
     \label{eq:loge} S_i^ e &= \frac{F_v^i\cdot E_t^e}{||F_v^i||\cdot ||E_t^e||}\\
     \label{eq:logp} S_i^ + &= \frac{F_v^i\cdot E_t^+}{||F_v^i||\cdot ||E_t^+||}\\
     \label{eq:logn} S_i^ - &= \frac{F_v^i\cdot E_t^-}{||F_v^i||\cdot ||E_t^-||}
\end{align}

In order to make a single prediction for the whole image, we aggregate the logit maps $S_{i}^+$ and  $S_{i}^-$  into $S^+$ and  $S^-$ according to the magnitude of $S_i^e$, and achieve the evidence-guided spatial aggregation:
\begin{align}
    \label{eq:scorep}\delta(E^I,E_t^e,E_t^+) &= \sum_i \big(\text{softmax}(S_{i}^e) \cdot S_{i}^+ \big) \\
    \label{eq:scoren} \delta(E^I,E_t^e,E_t^-) &= \sum_i \big(\text{softmax}(S_{i}^e) \cdot S_{i}^- \big) 
\end{align}
Then, the prediction for the target class can be made by applying Eq.~\ref{eq:prediction}. Notably, we do not introduce any new parameters in our re-formulation of the spatial aggregation function. All parameters used to project visual features to the textual space are inherited from the original multi-headed attention pooling layer in CLIP.

\begin{figure}[t]
\includegraphics[width=\linewidth]{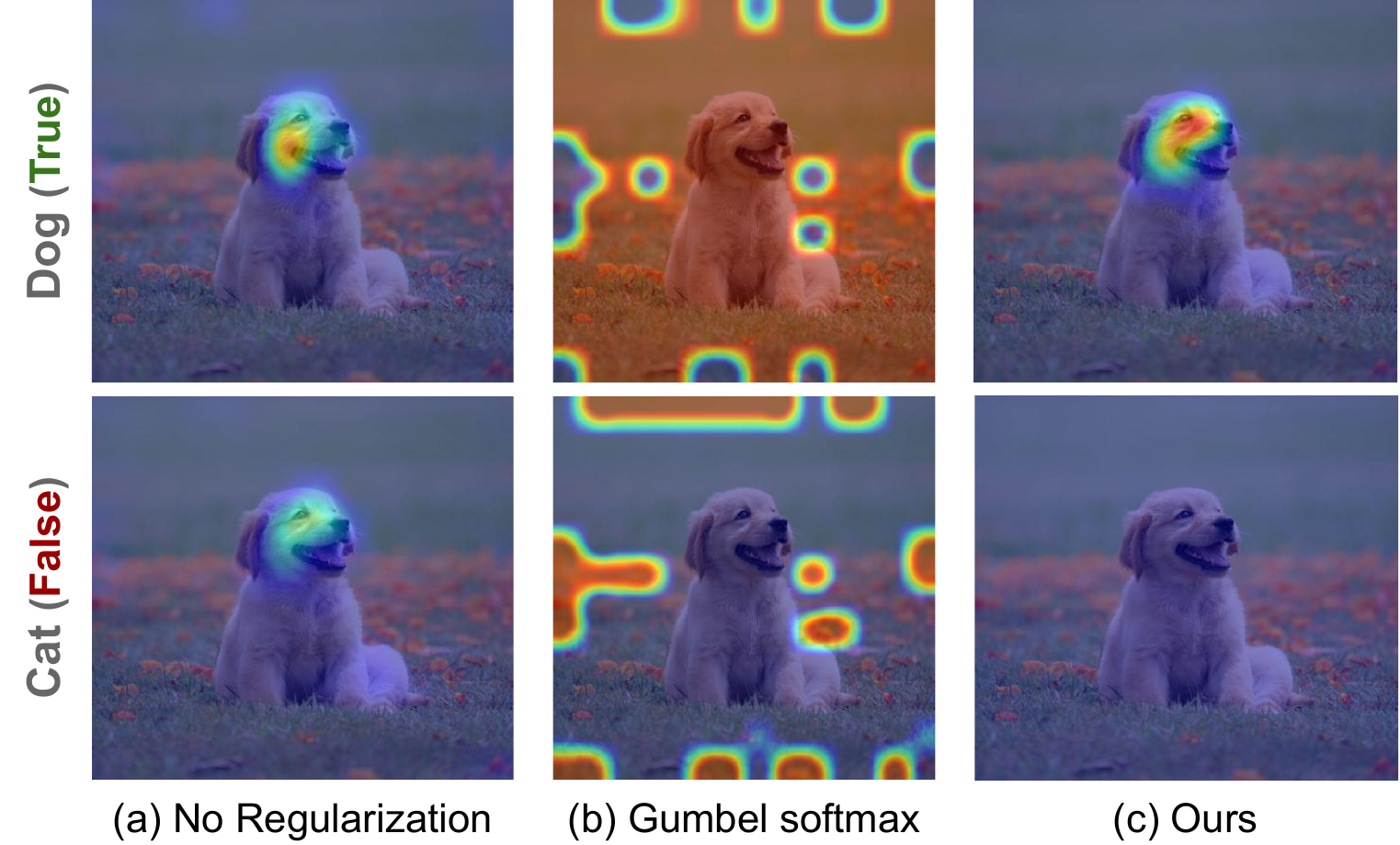}
\caption{\small 
    \textbf{Positive logit maps for different types of inter-class regularization. } (a) Without any regularization, the head area of the dog shows large positive logits to both ``Cat'' and ``Dog''. (b) Gumbel softmax always highlights the larger logit, correlating class labels to background regions. (c) Our proposed WTA module can highlight true-positive logits and suppress false-positive logits.
    }
   
\label{fig:WTA}
\end{figure}

\vspace{1mm}
\noindent\textbf{Winner-Take-All Regularization.} So far, the proposed framework learns to make predictions for different classes independently. Yet as shown in Fig.~\ref{fig:WTA} (a), treating classes independently can result in false-positive predictions, as the same visual features may positively respond to multiple similar classes, especially under the limited-annotation regime where sufficient supervision is lacking. We address this issue with a Winner-Take-All (WTA) module that regularizes each spatial region to only positively respond to at most one class. Given an image region $i$ and a label set containing $M$ target categories, we have $M$ positive logit scores $S_i^+ =[(S_i^+)^0, (S_i^+)^1,...,(S_i^+)^{M-1} ]$ based on Eq.~\ref{eq:logp}. The regularization weights $ w_i\in \mathcal{R}^M$ are computed over all the labels,
\begin{align}
    w_i= \text{softmax}\big(\gamma\cdot S_i^+\cdot \text{max$_m$}(S_i^+) \big)
\end{align}
where $\text{max$_m$}(S_i^+)$ represents the maximum logit score for regions $i$ across the $M$ classes, and $\gamma$ is a hyperparameter. 

Then, we update the positive logits elementwisely as,
\begin{align}
    (S_i^+)' = w_i\odot S_i^+
\end{align}
where $\odot$ is the Hadamard product. As illustrated in Fig.~\ref{fig:WTA} (b)(c), in contrast to Gumbel softmax which always signalizes the maximum element, our WTA  highlights the larger elements only if more than one logit has large values, which ensures that an image region can positively respond to none or just one of the given classes.

\vspace{1mm}
\noindent\textbf{Optimization}. 
We apply the Asymmetric Loss (ASL)~\cite{ridnik2021asymmetric} to handle the inherent positive-negative imbalance in the optimization of multi-label recognition. Specifically, we compute losses for a positive (image, label) pair  $\mathcal{L}_{+}$ and a negative (image, label) pair  $\mathcal{L}_{-}$ as follows:
\begin{align}
    \label{eq:aslp}&\mathcal{L}_{+} = (1-p)^{\gamma_{+}} \log(p) \\
    \label{eq:asln}& \mathcal{L}_{-} = (p_c)^{\gamma_{-}} \log (1 - p_c)
\end{align}
where $p$ is the probability in Eq.~\ref{eq:prediction}, and  $p_c = \max(p-c, 0)$ is the probability for negative examples shifted by hard thresholding via the margin $c$. We set the hyper-parameters $\gamma_{-} \ge \gamma_{+}$, so that ASL down-weighs
and hard-thresholds easy negative samples. The pair of learnable prompts are updated by back-propagating ASL through the frozen text encoder. 

\vspace{1mm}
\noindent\textbf{Methodology Rationale}. In this part, we delve deeper into the analysis to elucidate the advantages conferred by the negative prompts and evidential prompts. When minimizing the asymmetric loss (see Eq.\ref{eq:aslp} and\ref{eq:asln}), optimizing the positive prompts entails refining true positive predictions for true classes (i.e., classes present in the image) and true negative predictions for false classes (i.e., classes not present in the image). By incorporating negative prompts, the model untangles the optimization process to explicitly further refine false negative predictions for true classes and false positive predictions for false classes. This approach introduces independent and complementary contexts, thereby enhancing the model's discriminatory and generalization capabilities. Furthermore, this optimization strategy facilitates classification by comparing positive and negative logits, eliminating the need for manual threshold selection typically required in positive-only models. 

Building upon the aforementioned analysis, we introduce positive and negative prompts in \ourconf~\cite{sun2022dualcoop}, substantiating their efficacy through empirical results in Sec.\ref{sec:exp_analysis}. Adhering to the formulations in Eq.\ref{eq:scorep} and~\ref{eq:scoren}, the final positive/negative logits in \ourconf are expressed as follows:

\begin{align}
\label{eq:dcp}\delta^+ &= \sum_i \big(\text{softmax}(S_{i}^+) \cdot S_{i}^+ \big) \\
\label{eq:dcn}\delta^- &= \sum_i \big(\text{softmax}(S_{i}^+) \cdot S_{i}^- \big)
\end{align}
where $S_{i}^+$ and $S_{i}^-$ denote the positive and negative logit maps computed in Eq.\ref{eq:logp} and\ref{eq:logn}, respectively. The fundamental concept is to achieve a balanced integration of positive and negative prompts, leveraging their respective logit maps for comprehensive and effective representation.

While \ourconf has demonstrated effectiveness, a lingering issue still hampers its representation capacity. In the case of a negative (image, label) pair during training, the objective seeks to minimize $\delta^+$ while maximizing $\delta^-$. As outlined in Eq.~\ref{eq:dcp}, minimizing $\delta^+$ leads to a reduction in correlation between the target class and regions of confusing background classes in $softmax(S_i^+)$. However, these confusing regions are crucial in learning effective negative contexts with robust discriminatory capabilities. Consequently, a distracted aggregation map $softmax(S_i^+)$ in Eq.~\ref{eq:dcn} arising from the effort to minimize $\delta^+$ may overlook the most informative spatial regions. This oversight ultimately restricts the model's ability to learn negative prompts effectively, limiting its overall performance.

The evidential prompts are proposed to alleviate this limitation by disentangling the aggregation map from positive logits. By allowing the evidential logit map to capture all the correlated visual content to target classes, the positive and negative prompts can concentrate on recognizing the target classes and discerning confusing classes, respectively, without conflicts. As a result, this approach allows for more effective and conflict-free learning, and consistently yields improvements across various settings and datasets.

\begin{table*}
    \begin{center}
     \caption{\small \textbf{Multi-label Recognition on MS-COCO and VOC2007 with partial labels.} \ours achieves the best performance over all SOTA methods. $^*$ indicates previous models using weights pretrained by CLIP~\cite{radford2021learning}}~\label{table:partial_label}
        \resizebox{0.9\linewidth}{!}{
        \begin{tabular}{c| c| c c c c c c c c c | c }
            \Xhline{3\arrayrulewidth} 
            Methods & \cellcolor{yellow!15} \#P & $10\%$ & $20\%$  & $30\%$ & $40\%$ & $50\%$ & $60\%$ &   $70\%$ &  $80\%$ &  $90\%$  & \cellcolor{yellow!15}Avg. \\
            \Xhline{3\arrayrulewidth} 
            \multicolumn{12}{c}{MS-COCO~\cite{lin2014microsoft}} \\
            \Xhline{3\arrayrulewidth}
            SSGRL~\cite{chen2019learning} & \cellcolor{yellow!15} 64.7M & 62.5 & 70.5 & 73.2 & 74.5 & 76.3 & 76.5 &  77.1 & 77.9 & 78.4 & \cellcolor{yellow!15} 74.1  \\
            GCN-ML~\cite{chen2019multi} & \cellcolor{yellow!15} 44.9M &  63.8 & 70.9 & 72.8 & 74.0 & 76.7 & 77.1 & 77.3 & 78.3 & 78.6 & \cellcolor{yellow!15} 74.4 \\
            KGGR~\cite{chen2020knowledge} & \cellcolor{yellow!15} $\ge$ 25M  & 66.6 & 71.4 & 73.8 & 76.7 & 77.5 & 77.9 & 78.4 & 78.7 & 79.1 & \cellcolor{yellow!15} 75.6 \\
            Curriculum labeling~\cite{durand2019learning} & \cellcolor{yellow!15} $\ge$ 38M &  26.7 & 31.8 & 51.5 & 65.4 & 70.0 & 71.9 & 74.0 & 77.4 & 78.0 & \cellcolor{yellow!15} 60.7 \\
            Patial BCE~\cite{durand2019learning} & \cellcolor{yellow!15} $\ge$ 38M &  61.6 & 70.5 & 74.1 &  76.3 & 77.2 &  77.7 & 78.2 &  78.4 & 78.5 &\cellcolor{yellow!15} 74.7\\
            SST~\cite{chen2021structured} & \cellcolor{yellow!15} 33.5M  & 68.1 &  73.5 & 75.9 & 77.3 & 78.1 & 78.9 & 79.2 & 79.6 & 79.9 & \cellcolor{yellow!15} 76.7 \\
            SST$^*$ &  \cellcolor{yellow!15} 33.5M  &  69.1 & {78.5} & {79.3} & {79.9} & 80.1 & {80.5} & {81.1} & {80.7} & 80.7 & \cellcolor{yellow!15} 78.9 \\
            SARB~\cite{pu2022semantic} & \cellcolor{yellow!15} 29.6M &  71.2 & 75.0 & 77.1 & 78.3 & 78.9 & 79.6 & 79.8 &  80.5 & 80.5 & \cellcolor{yellow!15} 77.9 \\
            SARB$^*$ & \cellcolor{yellow!15} 29.6M & {75.5} & {78.5} & 79.0 & 79.5 & {80.4} & 80.2 & 80.8 & 80.6 & {80.8} & \cellcolor{yellow!15} {79.4} \\      
            SCPNet~\cite{ding2023exploring} & \cellcolor{yellow!15} {3.4M} & \underline{80.3} & \underline{82.2} & \underline{82.8} & \underline{83.4} & \underline{83.8} & \underline{83.9} & \underline{84.0} & \underline{84.1} & \underline{84.2} & \cellcolor{yellow!15} \underline{83.2} \\  
            \ourconf~\cite{sun2022dualcoop} & \cellcolor{yellow!15} \textbf{1.3M} &  {78.7} & {80.9} & {81.7} & {82.0} & {82.5} & {82.7} & {82.8} & {83.0} & {83.1} & \cellcolor{yellow!15} {81.9} \\      
            \ours  & \cellcolor{yellow!15} \underline{1.5M} &  \textbf{81.4} & \textbf{83.1} & \textbf{83.7} & \textbf{84.2} & \textbf{84.4} & \textbf{84.5} & \textbf{84.8} & \textbf{85.0} & \textbf{85.1} & \cellcolor{yellow!15} \textbf{84.0} \\       

            \Xhline{3\arrayrulewidth} 
              \multicolumn{12}{c}{PASCAL VOC 2007 ~\cite{everingham2010pascal}}\\
            \Xhline{3\arrayrulewidth}
             SSGRL~\cite{chen2019learning} & \cellcolor{yellow!15}66.6M  &77.7 & 87.6 & 89.9 & 90.7 & 91.4 & 91.8 & 91.9 & 92.2 & 92.2 & \cellcolor{yellow!15} 89.5\\
            GCN-ML~\cite{chen2019multi} & \cellcolor{yellow!15}44.9M &  74.5 & 87.4 & 89.7 & 90.7 & 91.0 & 91.3 & 91.5 & 91.8 & 92.0 & \cellcolor{yellow!15} 88.9\\
            KGGR~\cite{chen2020knowledge} & \cellcolor{yellow!15}$\ge$ 25M  & 81.3 & 88.1 & 89.9 & 90.4 & 91.2 & 91.3 & 91.5 & 91.6 & 91.8 & \cellcolor{yellow!15} 89.7 \\
            Curriculum labeling~\cite{durand2019learning} & \cellcolor{yellow!15} $\ge$ 38M  & 44.7 & 76.8 & 88.6 & 90.2 & 90.7 & 91.1 & 91.6 & 91.7 & 91.9 & \cellcolor{yellow!15} 84.1\\
            Patial BCE~\cite{durand2019learning} & \cellcolor{yellow!15} $\ge$ 38M &80.7 & 88.4 & 89.9 &  90.7 & 91.2 & 91.8 & 92.3 & 92.4 & 92.5  & \cellcolor{yellow!15} 90.0 \\
            SST~\cite{chen2021structured} & \cellcolor{yellow!15} 32.4M & 81.5 & 89.0 & 90.3 & 91.0 & 91.6 &  92.0 & 92.5 & 92.6 & 92.7 & \cellcolor{yellow!15} 90.4 \\
            SARB~\cite{pu2022semantic} & \cellcolor{yellow!15} 29.6M &  {83.5} & {88.6} & {90.7} & {91.4} & {91.9} & {92.2} & {92.6} & {92.8} & {92.9} & \cellcolor{yellow!15} {90.7} \\
            SPCNet~\cite{ding2023exploring} &\cellcolor{yellow!15} \textbf{--} &  \underline{91.1} & \underline{92.8} & \underline{93.5} & \underline{93.6} & \underline{93.8} & \underline{94.0} & \underline{94.1} & \underline{94.2} & \underline{94.3} & \cellcolor{yellow!15} \underline{93.5} \\ 
            \ourconf~\cite{sun2022dualcoop} &\cellcolor{yellow!15} \textbf{0.3M} &  {90.3} & {92.2} & {92.8} & {93.3} & {93.6} & {93.9} & {94.0} & {94.1} & {94.2} & \cellcolor{yellow!15} {93.2} \\ 
            \ours  &\cellcolor{yellow!15} \underline{0.4M} &\textbf{92.7} & \textbf{93.4} & \textbf{93.8} & \textbf{94.0} & \textbf{94.3} & \textbf{94.4} & \textbf{94.4} & \textbf{94.7} & \textbf{94.9} & \cellcolor{yellow!15} \textbf{94.1} \\ 
             \Xhline{3\arrayrulewidth} 
        \end{tabular}
        } 
    \end{center}
\vspace{-10pt}
\end{table*}
\section{Experiments}\label{sec:experiments}

In this section, we first report the performance on partial-label and zero-shot multi-label recognition benchmarks, then present experiments to analyze the proposed method.

\subsection{Multi-Label Recognition with Partial Labels}\label{sec:mlc_pl}

\textbf{Datasets.} 
We conduct experiments on MS-COCO~\cite{lin2014microsoft}, VOC2007~\cite{everingham2010pascal}, and BigEarth~\cite{bigearh} to evaluate multi-label recognition with partial labels. MS-COCO~\cite{lin2014microsoft} contains 80 common object categories, and we use the official \texttt{train2014} (82K images) and \texttt{val2014} (40K images) splits for training and the inference. VOC2007~\cite{everingham2010pascal} contains 20 object categories and we use the official \texttt{trainval} (5K images) and \texttt{test} (5K images) splits for training and test. 
Furthermore, since CLIP pretraining data is not publicly available and it is plausible that CLIP pretraining data covers many coarse and fine-grained visual domains since it performs well in the zero-shot evaluation for many downstream tasks, we also experiment on a Remote Sensing Image dataset BigEarth~\cite{bigearh}, whose domain is far from the domains of the datasets in the mainstream papers (i.e. PASCAL VOC, MS-COCO, and NUS-WIDE). 

To create the training set with partial labels, we follow the standard practice~\cite{chen2021structured,durand2019learning,pu2022semantic} to mask out labels from the fully annotated training set, and use the remaining labels for training. The proportion of kept labels varies from $10\%$ to $90\%$ as in previous works~\cite{chen2021structured,pu2022semantic}.

\vspace{1mm}
\noindent\textbf{Evaluation.} 
We report the mean average precision (mAP) for each proportion of labels available for optimization (from $10\%$ to $90\%$) and its average value for all proportions.
We count the learnable parameters (\#P) of each baseline and \ours to measure the complexity of optimization.

\vspace{1mm}
\noindent\textbf{Implementation.} For fair comparison, we adopt ResNet-101~\cite{he2016deep} as the visual encoder in all baselines and \ours with input resolution 448$\times$448,  and use the same Transformer~\cite{radford2019language,vaswani2017attention} in CLIP~\cite{radford2021learning} as the text encoder. Visual and text encoders are initialized from the CLIP pretrained model and kept frozen during optimization. For each class/label, we learn three independent context vectors with 12 context tokens (N = 12) to keep a similar size of parameters to \ourconf. Note that these context tokens are the only learnable parts of \ours. We use the SGD optimizer with an initial rate of 0.002 which is decayed by the cosine annealing rule. We train context vectors for 50 epochs with a batch-size 32/32/8 for MS-COCO/BigEarth/VOC2007, respectively. For ASL loss, we choose $\gamma_{+} = 1$, $\gamma_{-}=2$ and $c=0.05$ via validation. Training is done with one RTX A6000.

\vspace{1mm}
\noindent\textbf{Baselines.} To evaluate the effectiveness of \ours, we compare with the following baselines: 
\begin{itemize}
    \item SSGRL~\cite{chen2019learning}, GCN-ML~\cite{chen2019multi} and KGGR~\cite{chen2020knowledge} that adopt graph neural networks for label dependencies. 
    \item Curriculum labeling~\cite{durand2019learning} and  SST~\cite{chen2021structured} that generate pseudo labels for unknown labels.
    \item Partial BCE~\cite{durand2019learning} that uses a normalized BCE loss to better exploit partial labels.
    \item SARB~\cite{pu2022semantic} that blends category-specific representation across different images to transfer information.
    \item SCPNet~\cite{ding2023exploring}, TaI-DPT~\cite{guo2023texts}, and our DualCoop~\cite{sun2022dualcoop} that adopt the large-scale pretrained vision-language model CLIP.
\end{itemize}

\vspace{1mm}
\noindent\textbf{Results.} 
Table~\ref{table:partial_label} shows the comparison of mAP between \ours and all baselines optimized with $10\%$ to $90\%$ of labels. 
For the two recent works (SST~\cite{chen2021structured} and SARB~\cite{pu2022semantic}), we further substitute ImageNet pretrained weights~\cite{he2016deep} with CLIP pretrained weights~\cite{radford2021learning} when initializing their visual encoders, which results in $\text{SST}^*$ and $\text{SARB}^*$ in Table~\ref{table:partial_label}. 
Since we learn class-specific prompts, \ours on MS-COCO adopts more learnable parameters than VOC2007. 
The proposed \ours achieves the best performance across all proportions of labels available during training. Notably, \ours consistently improves \ourconf with similar learnable overhead, and outperforms the second-best method SCPNet~\cite{ding2023exploring} on both MS-COCO and VOC2007 with less than half of the learnable parameters. Especially, when only providing $10\%$ of labels during training, \ours improves \ourconf by more than 2$\%$ and outperforms SPCNet by more than 1$\%$ on both datasets. We also adopt the training protocols in TaI-DPT~\cite{guo2023texts}, which uses a larger batch size and more tunable hyperparameters, and compare the results in Table~\ref{table:tai}.  Without extra training data, \ours can further boost the performance and consistently outperform TaI-DPT, which exploits rich captioning data for training and two CLIP models for inference. This indicates \ours's ability to quickly adapt to the multi-label recognition task with a few labels. On BigEarth, we compare \ours with \ourconf and a strong baseline SARB. Table~\ref{table:bigearth} shows that \ours consistently improves over \ourconf and SARB with significant gaps under different portions of labels, proving that \ours boosts performance in various visual domains by taking advantage of the powerful vision-language pretraining.

\begin{table}[t]
    \begin{center}
     \caption{\small \textbf{Comparison between TaI-DPT~\cite{guo2023texts} and our  methods on MS-COCO.} All follow the same training setting~\cite{guo2023texts}.}
     \label{table:tai}
        \resizebox{\linewidth}{!}{
        \begin{tabular}{c | c | c c c c c | c}
            \Xhline{3\arrayrulewidth} 
            Method &\cellcolor{yellow!15}\#P &$10\%$  & $30\%$ & $50\%$ & $70\%$ & $90\%$  &\cellcolor{yellow!15} Avg.\\
            \Xhline{3\arrayrulewidth}
             {TaI-DPT}     &\cellcolor{yellow!15} {1.3M} &81.5   &83.3   &83.9   &84.2   &84.5   &\cellcolor{yellow!15}83.5\\ 
             {\ourconf} &\cellcolor{yellow!15} {\tbf{1.3M}}  &81.0   &82.9   &83.5   &84.0   &84.3   &\cellcolor{yellow!15}83.1\\ 
             {\ours}                           &\cellcolor{yellow!15} {1.5M}  &\tbf{81.9}   &\tbf{84.1}   &\tbf{84.6}   &\tbf{85.0}   &\tbf{85.4}   &\cellcolor{yellow!15}\tbf{84.2}\\   
             \Xhline{3\arrayrulewidth} 
        \end{tabular}
        } 
    \end{center}
\end{table}

\begin{table}[t]
    \begin{center}
     \caption{\small \textbf{Comparison between SARB~\cite{pu2022semantic} and our methods on BigEarth.}  All use parameters pretrained by CLIP~\cite{radford2021learning}.}
     \label{table:bigearth}
        \resizebox{\linewidth}{!}{
        \begin{tabular}{c | c |c c c c c | c}
            \Xhline{3\arrayrulewidth} 
            Method &\cellcolor{yellow!15}\#P & $10\%$  & $30\%$ & $50\%$ & $70\%$ & $90\%$  &\cellcolor{yellow!15} Avg.\\
            \Xhline{3\arrayrulewidth}  
            {SARB}     &\cellcolor{yellow!15} {29.6M} &71.5 &76.5 &78.6 &80.4 &84.2 &\cellcolor{yellow!15}78.2 \\
            \ourconf &\cellcolor{yellow!15}  \tbf{0.3M} &81.7 &86.5 &90.1 &91.7 &92.2 &\cellcolor{yellow!15}88.4 \\
            \ours                           &\cellcolor{yellow!15}  0.4M &\tbf{83.4} &\tbf{90.3} &\tbf{91.9} &\tbf{92.3} &\tbf{93.0} &\cellcolor{yellow!15}\tbf{90.1} \\
             \Xhline{3\arrayrulewidth} 
        \end{tabular}
        } 
    \end{center}
\end{table}

\begin{table}[t]
    \begin{center}
     \caption{\small \textbf{Computations Efficiency Comparison. } }
     \label{table:computation_cost}
        \begin{tabular}{ccccc}
            \Xhline{3\arrayrulewidth} 
            &\multicolumn{2}{c}{Training}            &\multicolumn{2}{c}{Testing}\\
            \cmidrule(lr){2-3} \cmidrule(lr){4-5}
             \multirow{2}{*}{\makecell{Methods}} &    Latency    & Memory  & Latency  & Memory  \\ 
              &  ms/img    &  GB/img  &   ms/img   & GB/img  \\ 
            \Xhline{3\arrayrulewidth} 
           SARB\cite{pu2022semantic} & 4.7 & 0.21 & 4.0 & 0.13\\
           TaI-DPT~\cite{guo2023texts} & -- & -- & 4.8 & 0.09\\
           \ourconf~\cite{sun2022dualcoop} & 5.3 & 0.22 & 4.0 & 0.06\\
           \ours & 6.6 & 0.22 & 4.0 & 0.06\\
             \Xhline{3\arrayrulewidth} 
        \end{tabular}
    \end{center}
\end{table}

\begin{table*}
    \begin{center}
     \caption{\small \textbf{Zero-Shot Multi-label Recognition on NUS-WIDE.} \ours achieves the best F1 score over all SOTA methods at Top-3/Top-5 predictions in both ZSL and GZSL settings.  }
     \label{table:zsl_nus_wide}
        \resizebox{\linewidth}{!}{
        \begin{tabular}{c c c c c c c c c c c c c c c c}
        \Xhline{3\arrayrulewidth} 
        && \multicolumn{7}{c}{Zero-Shot Learning (ZSL)} &\multicolumn{7}{c}{Zero-Shot Learning (GZSL)} \\
        \cmidrule(lr){3-9} \cmidrule(lr){10-16}
        \multirow{2}{*}{Methods} & \cellcolor{yellow!15} & \multicolumn{3}{c}{Top-3} & \multicolumn{3}{c}{Top-5} & \cellcolor{yellow!15} & \multicolumn{3}{c}{Top-3} & \multicolumn{3}{c}{Top-5} & \cellcolor{yellow!15}  \\
        & \cellcolor{yellow!15} \multirow{-2}{*}{\#P} & \textbf{P} & \textbf{R} &  \textbf{F1} & \textbf{P} & \textbf{R} & \textbf{F1} & \cellcolor{yellow!15} \multirow{-2}{*}{mAP}& \textbf{P} & \textbf{R} & \textbf{F1} & \textbf{P} & \textbf{R} & \textbf{F1} & \cellcolor{yellow!15} \multirow{-2}{*}{mAP}\\
        \Xhline{3\arrayrulewidth} 
CONSE~\cite{norouzi2013zero}            & \cellcolor{yellow!15} -           & 17.5      & 28.0      & 21.6 & 13.9 & 37.0  &  20.2    &\cellcolor{yellow!15} 9.4 & 11.5 & 5.1    &7.0  & 9.6   & 7.1   &8.1      &  \cellcolor{yellow!15}  2.1  \\ 
LabelEM~\cite{akata2015label}           & \cellcolor{yellow!15} -           & 15.6      & 25.0      & 19.2 & 13.4 & 35.7  &  19.5    &\cellcolor{yellow!15} 7.1 & 15.5 & 6.8    &9.5  & 13.4  & 9.8   & 11.3     &\cellcolor{yellow!15}  2.2  \\ 
Fast0Tag~\cite{zhang2016fast}           & \cellcolor{yellow!15} 0.61M       &  22.6     & 36.2      & 27.8 & 18.2 & 48.4  & 26.4    &\cellcolor{yellow!15} 15.1 & 18.8 & 8.3    &11.5 & 15.9  & 11.7  & 13.5     & \cellcolor{yellow!15}  3.7\\
OAL~\cite{Kim2018}                      & $\cellcolor{yellow!15} \ge$ 12.8M &  20.9     & 33.5      & 25.8 & 16.2 & 43.2  & 23.6    &\cellcolor{yellow!15} 10.4 & 17.9 & 7.9    &10.9 & 15.6  & 11.5  & 13.2     & \cellcolor{yellow!15}  3.7\\
LESA$_{\text{M10}}$~\cite{huynh2020shared}& \cellcolor{yellow!15} $\ge$0.45M&  {25.7}   & 41.1      & 31.6 &{19.7}& 52.5  & 28.7    &\cellcolor{yellow!15} 19.4 & 23.6 & 10.4   & 14.4 & 19.8  & 14.6  & 16.8     &\cellcolor{yellow!15}   5.6 \\
BiAM~\cite{narayan2021discriminative}   & \cellcolor{yellow!15} 3.8M        & --        & --        & {33.1}&   -- & --    &  {30.7}  &\cellcolor{yellow!15} {26.3}  & --   & --     &  16.1& --    & --    & 19.0     & \cellcolor{yellow!15} 9.3 \\
SDL$_{\text{M7}}$~\cite{ben2021semantic}& \cellcolor{yellow!15} 33.6M       & 24.2      &{41.3}     & 30.5 & 18.8 & {53.4}&  27.8    &\cellcolor{yellow!15} 25.9  &{27.7}&\udl{13.9} &{18.5}& {23.0}&\udl{19.3} & {21.0}   & \cellcolor{yellow!15} \udl{ 12.1} \\
\ourconf~\cite{sun2022dualcoop}         & \cellcolor{yellow!15} \tbf{0.07M} &\udl{37.3} &\udl{46.2} & \udl{41.3}&\udl{28.7}& \udl{59.3}&  \udl{38.7}  & \cellcolor{yellow!15} \udl{43.6}  &\udl{31.9}&\udl{13.9} &\udl{19.4}&\udl{26.2}& 19.1  & \udl{ 22.1}  & \cellcolor{yellow!15} {12.0} \\
\ours                                   & \cellcolor{yellow!15} \tbf{0.07M} & \tbf{42.4}&\tbf{52.5} & \tbf{46.9}&\tbf{31.2}& \tbf{64.5}&  \tbf{42.1}  & \cellcolor{yellow!15} \tbf{47.1}  &\tbf{34.7}& \tbf{15.2} &\tbf{21.1}& \tbf{29.2}& \tbf{21.3}  &\tbf{ 24.7}  & \cellcolor{yellow!15} \tbf{15.1} \\
\Xhline{3\arrayrulewidth} 
    \end{tabular}
    } 
    \end{center}
\end{table*}

\vspace{1mm}
\noindent\textbf{Full Label Training.} 
On MS-COCO,  we also learn with 100\% of training labels. Without fine-tuning the visual encoder, \ours achieves 85.3\% mAP, which improves \ourconf by 2$\%$ and outperforms previous SOTA approaches like ASL~\cite{ridnik2021asymmetric}~(85.0\% mAP) and CSRA~\cite{zhu2021residual}~(83.5\% mAP) with the same ResNet-101 backbone. This shows \ours's promising ability to exploit the pretrained CLIP model for addressing challenging MLR tasks.

\vspace{1mm}
\noindent\textbf{Computational Cost.} 
We compare the computational cost between \ours and previous methods in terms of training/testing latency and memory (see Table~\ref{table:computation_cost}) using the same device (one Nividia A100 GPU). For the current multi-label recognition task, the categories are pre-set before inference (i.e. we already know which class we would like to consider during inference.) In this case, we compute the text features for each class from the learned prompts and the class name prior to inference. Then we use the pre-computed text features to predict each image during the test. Since the text features are pre-computed (very lightweight computing overhead), the text encoder is not executed during inference. For inference, the latency time and memory consumption of \ours are the same as \ourconf when using the same backbone for the image encoder. During training, CLIP-based methods slightly raise latency time and memory consumption, since image and text encoders are both executed during the forward, and only prompts are updated in \ours.

\subsection{Zero-shot Multi-Label Recognition}
\vspace{1mm}
\textbf{Datasets.} 
Following \cite{ben2021semantic, huynh2020shared}, we conduct experiments on MS-COCO~\cite{lin2014microsoft} and NUS-WIDE~\cite{chua2009nus} to perform zero-shot multi-label recognition. On MS-COCO, we follow \cite{bansal2018zero, ben2021semantic} to split the dataset into 48 seen classes and 17 unseen classes. NUS-WIDE~\cite{chua2009nus} dataset includes 270K images. Following \cite{ben2021semantic, huynh2020shared} we use 81 human-annotated categories as unseen classes and an additional set of 925 labels obtained from Flickr tags as seen classes.

\vspace{1mm}
\noindent\textbf{Evaluation.}
We follow \cite{ben2021semantic} and report precision, recall, and F1 score at Top-3 predictions in each image on MS-COCO. We also follow \cite{ben2021semantic, huynh2020shared} to report mAP over all categories as well as precision, recall, and F1 score in the Top-3 and Top-5 predictions in each image on NUS-WIDE. We evaluate all methods with both the zero-shot setting (test only on unseen classes) and the generalized zero-shot setting (test on both seen and unseen classes).

\vspace{1mm}
\noindent\textbf{Implementation.} We adopt ResNet-50~\cite{he2016deep} similar to \cite{ben2021semantic} as the visual encoder in \ours for input resolution 224. 
Instead of learning class-specific prompts, we learn the class-agnostic context vectors with 42 context tokens (N = 42) for all classes, which is the only learnable part in \ours.  
We optimize context vectors for 50 epochs with a batch size of 32/192 for MS-COCO/NUS-WIDE, respectively. During inference, we combine the learned pair of context vectors with the class name for each class (either base class or novel class) and compute the text features. Other implementation details are the same as in Sec.~\ref{sec:mlc_pl}

\vspace{1mm}
\noindent\textbf{Baselines.}
To evaluate the effectiveness of \ours in the zero-shot setting, we compare with the following baselines: 
\begin{itemize}
    \item CONSE~\cite{norouzi2013zero} that adopts an ensemble of classifiers for unseen classes.
    \item LabelEM~\cite{akata2015label} that learns a joint image-label embedding.
    \item Fast0Tag~\cite{zhang2016fast} and SDL~\cite{ben2021semantic} that estimate one or multiple diverse principal directions of the input images.
    \item Deep0Tag~\cite{rahman2018deep} and LESA~\cite{huynh2020shared} that estimate the relevant regions via region proposals and attention techniques, respectively.
    \item BiAM~\cite{narayan2021discriminative} that enhances the region-based features to minimize inter-class feature entanglement.
    \item \ourconf~\cite{sun2022dualcoop} that is based on the pretrained CLIP model.
\end{itemize}

\begin{table}[t]
    \begin{center}
     \caption{\small{\textbf{Zero-Shot Multi-Label Recognition on MS-COCO}. \ours achieves the best F1 score in both ZSL and GZSL settings.}}~\label{tab:zero_shot_mscoco}
    
        \resizebox{0.95\linewidth}{!}{
        \begin{tabular}{c  c c cc c c c}
        \Xhline{3\arrayrulewidth} 
        \multirow{2}{*}{Methods}  & \multicolumn{3}{c}{ZSL}& & \multicolumn{3}{c}{GZSL} \\
        \cmidrule(lr){2-4} \cmidrule(lr){6-8}
         & \textbf{P} & \textbf{R} & \cellcolor{yellow!15} \textbf{F1}& & \textbf{P} & \textbf{R} &\cellcolor{yellow!15} \textbf{F1} \\ 
           \Xhline{3\arrayrulewidth} 
           CONSE~\cite{norouzi2013zero}     & 11.4       & 28.3  &  \cellcolor{yellow!15}  16.2 & & 23.8 & 28.8 &  \cellcolor{yellow!15} 26.1 \\
           Fast0Tag~\cite{zhang2016fast}    & 24.7       & 61.4  &  \cellcolor{yellow!15}  25.3 && 38.5 & 46.5 &  \cellcolor{yellow!15}  42.1 \\
           Deep0Tag~\cite{rahman2018deep}   & {26.5}     & {65.9}&  \cellcolor{yellow!15} {37.8}& &  43.2 &  52.2 &  \cellcolor{yellow!15}  47.3 \\
           SDL$_{\text{M2}}$~\cite{ben2021semantic} & 26.3       & 65.3  &  \cellcolor{yellow!15}  37.5 &&  {59.0} & {60.8} &  \cellcolor{yellow!15}  {59.9} \\
           \hline
           CLIP~\cite{radford2021learning}  & {25.6}     & {63.6}& \cellcolor{yellow!15} {36.5} & &	{31.0} & {36.2} & \cellcolor{yellow!15}  {33.4}	 \\
           CoOp~\cite{zhou2022conditional}  & {33.7}     & {83.8}& \cellcolor{yellow!15} {48.1} & &	{53.9} & {62.9} & \cellcolor{yellow!15}  {58.1}	 \\
           \ourconf~\cite{sun2022dualcoop}  & {35.3}     & {87.6}& \cellcolor{yellow!15} {50.3} & &	{58.4} & {68.1} & \cellcolor{yellow!15}  {62.9}	 \\
           \ours                            & \tbf{36.8} & \tbf{91.4} & \cellcolor{yellow!15} \tbf{52.5} & &	\tbf{59.4} & \tbf{69.3} & \cellcolor{yellow!15}  \tbf{64.0}	 \\
            \Xhline{3\arrayrulewidth} 
        \end{tabular}
        } 
    \end{center}
\end{table}

\vspace{1mm}
\noindent\textbf{Results.} 
Table~\ref{table:zsl_nus_wide}  and \ref{tab:zero_shot_mscoco} show the comparison between \ours and all SOTA methods of zero-shot learning and generalized zero-shot learning on NUS-WIDE and MS-COCO datasets. \ours achieves the best accuracy in all cases with a very light learnable overhead (0.07M) and improves the performance of zero-shot learning with significant margins.
Compared to the previous state-of-the-art SDL~\cite{ben2021semantic}, \ours improves ZSL performance by 15.0 @Top-3 on MS-COCO, and by 14.1 @Top-3 and 14.3 @Top-5 on NUS-WIDE.  This shows the power of exploiting the pretrained alignment of textual and visual spaces in CLIP via \ours to solve multi-label recognition. \ours also consistently improves the precision and recall of \ourconf in all settings, demonstrating the effectiveness of our proposed methods to suppress false predictions.

\begin{table}[h]
    \begin{center}
     \caption{\small \textbf{Comparison among methods on MS-COCO using partial labels with the same initialization.}}~\label{table:semantic_guide}
        \resizebox{0.99\linewidth}{!}{
        \begin{tabular}{c | c |c c c c c c }
            \Xhline{3\arrayrulewidth} 
             \multirow{2}{*}{\makecell{Methods}} &  \multirow{2}{*}{\makecell{ CLIP-Based}} & \multirow{2}{*}{\makecell{$10\%$}}  & \multirow{2}{*}{\makecell{$30\%$}} & \multirow{2}{*}{\makecell{$50\%$}} & \multirow{2}{*}{\makecell{$70\%$}} & \multirow{2}{*}{\makecell{$90\%$}} & \multirow{2}{*}{\makecell{Avg.}} \\
             &&\\
            \Xhline{3\arrayrulewidth}  
            SST~\cite{chen2021structured} & \xmark &  68.1 & 75.9 & 78.1 & 79.2 & 79.9 &76.2\\
            SARB~\cite{pu2022semantic} & \xmark &  71.2 & 77.1 & 78.9 & 79.8 & 80.5 &77.5\\
            \hline
            Disc. Label & \cmark & 70.6 & 75.1 &  76.5 & 77.3 & 78.0 &75.5\\
            SST~\cite{chen2021structured} & \cmark &  69.1 & 79.3 & 80.1 & 81.1 & 80.7 &78.1\\
            SARB~\cite{pu2022semantic} & \cmark &  75.5 & 79.0 & 80.4 & 80.8 & 80.8 &79.3\\
            \hline
            CLIP~\cite{radford2021learning} & \cmark & -- & -- & -- & -- & -- &51.0\\
            CoOp~\cite{zhou2022conditional} & \cmark & 63.0 & 68.5 & 69.2 & 71.5 & 75.0 &69.4\\
            \ourconf & \cmark & {78.7} &{ 81.7} & {82.5} & {82.8}  & {83.1 } &81.8\\
            \ours & \cmark & \textbf{81.4} &\textbf{ 83.7} & \textbf{84.4} & \textbf{84.8}  & \textbf{85.1 } &83.9\\
             \Xhline{3\arrayrulewidth} 
        \end{tabular}
        } 
    \end{center}
\end{table}

\subsection{Method Analysis}\label{sec:exp_analysis}
In this subsection, we first demonstrate the overarching concepts in this work and subsequently validate the detailed model designs with extensive experiments.

\vspace{1mm}
\noindent\textbf{Why does the proposed method work for MLR?} 
Despite the strong generalization capabilities demonstrated by the CLIP~\cite{radford2021learning} model across diverse concepts, addressing downstream tasks like MLR is still non-trivial.
As evident from Table~\ref{tab:zero_shot_mscoco} and~\ref{table:semantic_guide}, directly applying CLIP to MLR results in even worse performance compared to non-CLIP-based methods (e.g., CLIP vs. SDL in Table~\ref{tab:zero_shot_mscoco} and CLIP vs. SARB in Table~\ref{table:semantic_guide}). 
The observed performance gaps primarily stem from the domain gap and the misalignment of objectives between CLIP and the downstream tasks.
To tackle the domain gap challenge, we initially address it through prompt tuning. As illustrated in Table~\ref{tab:zero_shot_mscoco} and~\ref{table:semantic_guide}, the CoOp method significantly enhances performance by introducing learnable prompts into CLIP. However, it still falls short of surpassing non-CLIP methods like SDL and SARB in low-label MLR. This can be attributed to the unbalanced representation and optimization resulting from simply tuning positive prompts, leading to limited performance when generalizing to testing data. To overcome this, we introduce negative prompts and evidential prompts to explicitly regulate the tuning process and introduce complementary contexts, thereby enhancing the model's generalization ability, as demonstrated in Table~\ref{table:dqp_vs_dq_pl}, \ref{table:dqp_vs_dq_zsl}, and\ref{table:ablation_prompt}.
The second challenge limiting CLIP's MLR performance stems from its design for one-to-one visual-textual contrastive objectives, making it challenging to extract fine-grained spatial information crucial for MLR. This limitation is addressed by devising the regional feature aggregation and the Winner-Take-All regularization to discriminate spatial details, resulting in improved accuracy, as shown in Table~\ref{table:dqp_vs_dq_pl}, \ref{table:dqp_vs_dq_zsl}, and~\ref{table:mha_vs_conv_proj}. With these designs, the proposed method alleviates limitations and challenges in exploiting the rich semantics in CLIP to address MLR, achieving state-of-the-art performance across different settings and datasets.

\vspace{1mm}
\noindent\textbf{Effectiveness of Text Supervision.} To show the effectiveness of text supervision from label space, we compare the model learned with discrete label space (``Discrete Label'') between five methods (SST~\cite{chen2021structured}, SARB~\cite{pu2022semantic}, CoOp~\cite{zhou2022conditional}, \ourconf,  and  \ours) which introduce the textual space to utilize the contextual correlation of labels in Table~\ref{table:semantic_guide}. We find that methods with text supervision usually perform better than the method that only uses discrete labels. However, when the semantic annotations are limited, text supervision sometimes yields worse performance (e.g. mAP of SST is $1.5\%$ lower than Discrete Labels with only $10\%$ of labels). CoOp~\cite{zhou2022conditional} utilizes the visual-textual alignment. However, with the original multi-head attention and single positive prompt, it yields worse performance than Discrete Labels.
To better utilize well-trained alignment for MLR tasks, \ours learns a context triplet and adopts evidence-guided region feature aggregation, which leads to great performance (\textit{e.g.} $10.8\%$ higher than Discrete Labels with $10\%$ of labels) and quickly adapts to the dataset even with limited labels.

\begin{table}[t]
    \begin{center}
     \caption{\small \textbf{Partial-label MLR performance with 50\% annotations.} ``Evi.'' represents the evidence-guided spatial aggregation. ``WTA'' denotes the winner-take-all module in training.}
     \label{table:dqp_vs_dq_pl}
        \resizebox{\linewidth}{!}{
        \begin{tabular}{  c| c c| c c c c c c }
            \Xhline{3\arrayrulewidth}  
            Dataset    &Evi.  &WTA   &C\_P   &C\_R   &C\_F &O\_P   &O\_R   &O\_F \\
            \Xhline{1.5\arrayrulewidth}
            \multirow{ 3}{*}{MSCOCO} & &  &72.1	 &80.4	 &75.8	 &73.7	 &83.9	 &78.5\\
             &\cmark &                    &74.3	&80.3	&77.0	&76.1	&83.7	&79.7\\
             &\cmark &\cmark              &\tbf{76.0}	&\tbf{80.9}	&\tbf{77.9}	&\tbf{77.0}	&\tbf{84.1}	&\tbf{80.4}\\
            \Xhline{1.5\arrayrulewidth} 
            \multirow{ 3}{*}{VOC} & &   &80.6	&93.4	&86.3	&82.4	&94.0	&87.8\\
             &\cmark &                  &81.8	&93.1	&86.4	&83.1	&93.9	&88.2\\
             &\cmark &\cmark            &\tbf{82.6}	&\tbf{93.5}	&\tbf{87.5}	&\tbf{84.2}	&\tbf{94.1}	&\tbf{88.9}\\
            \Xhline{3\arrayrulewidth} 
        \end{tabular}
        } 
    \end{center}
\end{table}

\begin{table}[t]
    \begin{center}
     \caption{\small{\textbf{Zero-shot MLR performance.} ``Evi.'' represents the evidence-guided spatial aggregation. ``WTA'' denotes the winner-take-all module in training.}}
     \label{table:dqp_vs_dq_zsl}
        \resizebox{\linewidth}{!}{
        \begin{tabular}{c | c c |c c c| c  c c}
        \Xhline{3\arrayrulewidth} 
            & &   & \multicolumn{3}{c|}{ZSL} & \multicolumn{3}{c}{GZSL} \\
            &Evi. &WTA & \textbf{P} & \textbf{R} &  \textbf{F1} & \textbf{P} & \textbf{R} &\textbf{F1} \\ 
           \Xhline{1.5\arrayrulewidth} 
        \multirow{3}{*}{MS-COCO}&&              &35.3	&87.6	&50.3		&58.4	&68.1	&62.9	\\
                                &\cmark&        &35.9	&89.2	&51.2		&59.0	&68.9	&63.6	\\
                                &\cmark&\cmark  &\tbf{36.8}	&\tbf{91.4}	&\tbf{52.5}		&\tbf{59.4}	&\tbf{69.3}	&\tbf{64.0}	\\
            \Xhline{1.5\arrayrulewidth} 
        \multirow{3}{*}{NUS-WIDE}&&      		&37.3	&46.2	&41.3		&31.9	&13.9	&19.4	\\
                                &\cmark&        &37.9	&46.3	&41.7		&32.8	&14.3	&19.9	\\
                                &\cmark&\cmark  &\tbf{42.4}	&\tbf{52.5}	&\tbf{46.9}		&\tbf{34.7}	&\tbf{15.2}	&\tbf{21.1}	\\
            \Xhline{3\arrayrulewidth} 
        \end{tabular}
        } 
    \end{center}
\end{table}

\vspace{1mm}
\noindent\textbf{CoOp v.s. \ours v.s. \ourconf.}
In Table.\ref{tab:zero_shot_mscoco} and\ref{table:semantic_guide}, we conduct a performance comparison between CoOp and the proposed methods. In the context of zero-shot MLR, both \ourconf and \ours exhibit notable improvements in the F1 metric. Specifically, \ourconf achieves a more than $2\%$ enhancement, while \ours outperforms by more than $4\%$ for both zero-shot recognition and generalized zero-shot recognition tasks. In partial-label settings, our methods consistently elevate CoOp's performance by more than $8\%$ across various proportions of available labels. This consistent improvement underscores the effectiveness of our proposed methods in recognizing multiple labels from images.
We also analyze the impact of the newly introduced components in \ours in Table.~\ref{table:dqp_vs_dq_pl} and~\ref{table:dqp_vs_dq_zsl}. For partial-label recognition, we report the per-class and the average overall precision (C\_P and O\_P), recall (C\_R and O\_R), and F1 measure (C\_F and O\_F). As shown in Table~\ref{table:dqp_vs_dq_pl}, introducing the evidence-guided spatial aggregation (denoted as ``Evi.'') significantly improves precision and the F measure, and further application of the winner-take-all module can further boost the performance. This demonstrates that our method can effectively relieve the limitation of \ourconf and suppress false positive predictions. For zero-shot recognition as shown in Tab.~\ref{table:dqp_vs_dq_zsl}, our proposed components show consistent improvements for all the metrics and datasets, demonstrating the ability to generalize to unseen classes. 

\begin{table}
    \begin{center}
     \caption{\small \textbf{Ablation on Linguistic Inputs for Zero-Shot Learning of MS-COCO.} M2$\sim$M5 have similar sizes of learnable parameters. }~\label{table:ablation_prompt}
        \resizebox{\linewidth}{!}{
        \begin{tabular}{c c c c c  c c c  }
            \Xhline{3\arrayrulewidth} 
            & \multirow{2}{*}{Linguistic Input}   & \multicolumn{3}{c}{ZSL} & \multicolumn{3}{c}{GZSL}  \\
            &  & \textbf{P} & \textbf{R} & \textbf{F1} & \textbf{P} & \textbf{R} & \textbf{F1} \\
            \Xhline{3\arrayrulewidth} 
            {M0} &Contextless               &  5.2 & 12.9 &  7.4 &  3.5 &  4.1 & 3.8\\
            {M1} &Hand-crafted              & 25.6 & 63.6 & 36.5 & 31.0 & 36.2 & 33.4\\
            {M2} &Neg. Prompt               &  9.5 & 23.1 & 13.5 & 2.5 &  3.3 & 2.8\\
            {M3} &Pos. Prompt               & 32.1 & 77.8 & 45.4 & 56.2 & 65.3 & 60.4\\
            {M4} &Pos.+Neg. Prompts         & 35.3 & 87.6 & 50.3 & 58.4 & 68.1 & 62.9\\
            {M5} &Evi.+Pos.+Neg. Prompts    & \tbf{35.9} & \tbf{89.2} & \tbf{51.2} & \tbf{59.0} & \tbf{68.9} &\tbf{63.6}\\
             \Xhline{3\arrayrulewidth} 
        \end{tabular}
        } 
    \end{center}
\end{table}

\begin{figure}[b]
	\centering
        \begin{minipage}[]{0.24\textwidth}
            \includegraphics[width=1\textwidth]{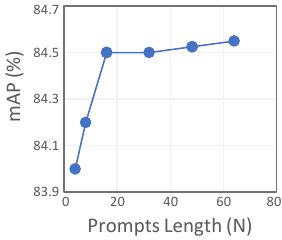}
             \vspace{-0.6cm}
             \subcaption{Partial-label MLR}
              \label{fig:abl_len1}
        \end{minipage}
        \begin{minipage}[]{0.24\textwidth}
            \includegraphics[width=1\textwidth]{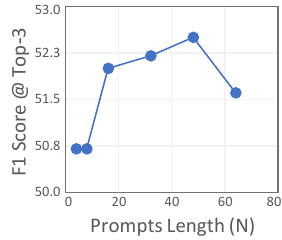}
             \vspace{-0.6cm}
             \subcaption{Zero-shot MLR}
              \label{fig:abl_len2}
        \end{minipage}
        \vspace{5mm}
        \begin{minipage}[]{0.24\textwidth}
            \includegraphics[width=1\textwidth]{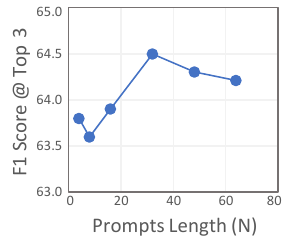}
             \vspace{-0.6cm}
             \subcaption{Generalized Zero-shot MLR}
              \label{fig:abl_len3}
        \end{minipage}
        \begin{minipage}[]{0.24\textwidth}
            \includegraphics[width=1\textwidth]{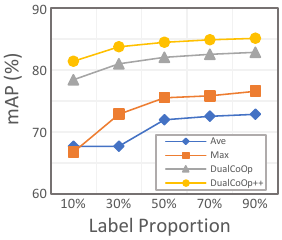}
             \vspace{-0.6cm}
             \subcaption{Partial-label MLR. }
              \label{fig:abl_aggre}
        \end{minipage}
	\caption{Analysis of different design options on MS-COCO.}
	\label{fig:ablation}
\end{figure}
\begin{figure*}
\begin{center}
     \includegraphics[width=0.98\linewidth]{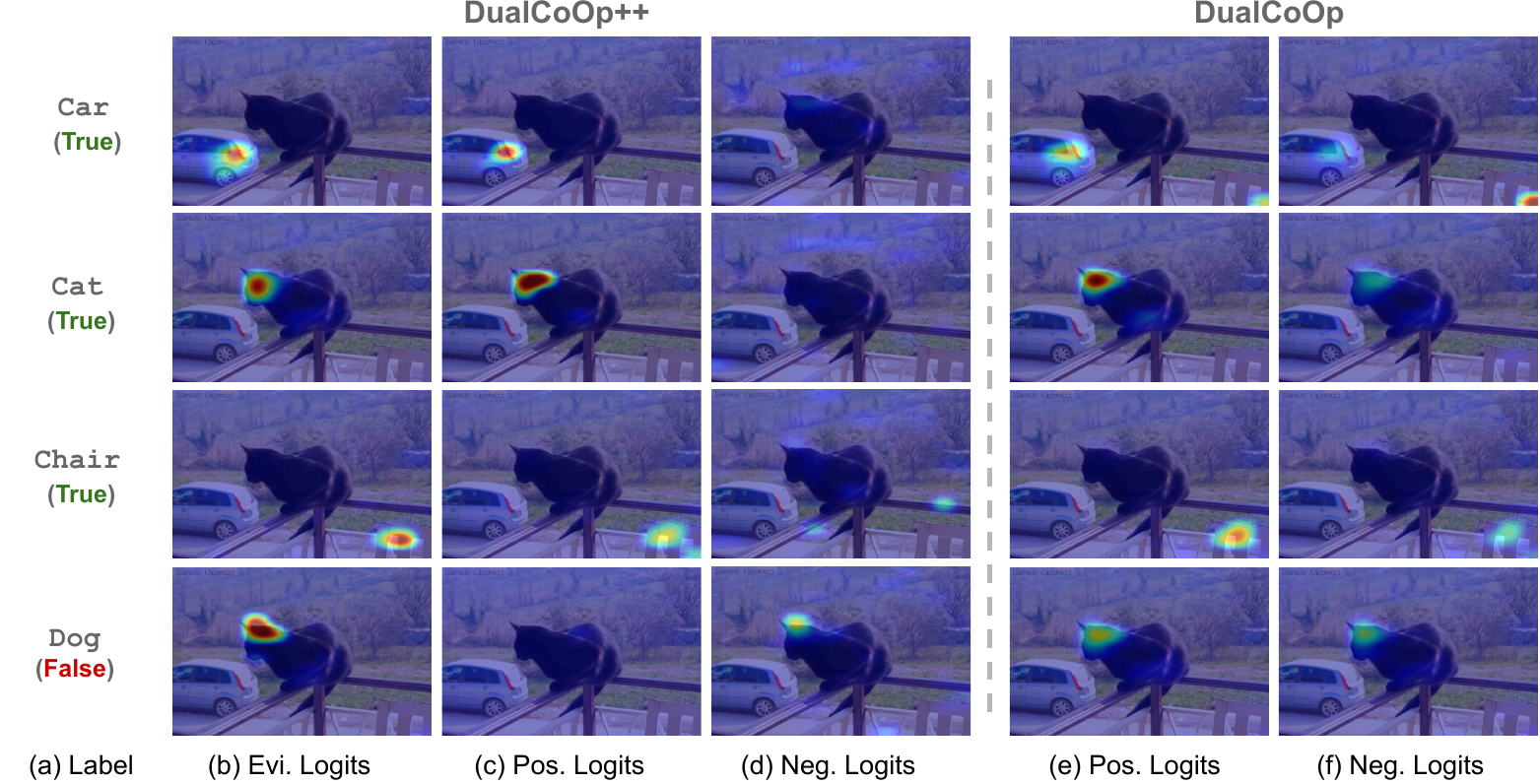}
\end{center} 
   \caption{\small 
   \textbf{Visualization of logit maps in \ours and \ourconf }. Given class labels (a) to \ours, the evidential logit map (b) highlights closely related image regions, and the positive (c)/ negative (d) logit maps provided correct positive and negative support for the highlighted areas to make final predictions. While in \ourconf, negative samples are not well optimized hence leading to weak negative logit maps, resulting in false-positive prediction (i.e. higher positive response than the negative response in the cat region for the label Dog ).
   }
   \label{fig:result}
   \vspace{-10pt}
\end{figure*}
\vspace{1mm}
\noindent\textbf{Ablation of Prompt Design.}  
We compare our proposed triple learnable prompts with hand-crafted prompts and single prompt learning method on the MS-COCO dataset with the zero-shot setting (see Table~\ref{table:ablation_prompt}).
Hand-crafted prompts can use either contextless class names~\cite{li2017learning} or manually designed prompt templates. In our experiments, we carefully chose the positive and negative prompt templates as ``a photo of a [classname]'' and ``a photo without a [classname]''. In contrast to performing the binary classification for each class with multiple learnable prompts as input, we also perform experiments with learning a single prompt of positive or negative contexts and use a chosen threshold (0.5 in our experiment) to make the prediction for each class. 
As we can see, the single positive prompt learning method (M3) performs better than non-learnable methods (M0 and M1), and a single negative learnable prompt (M2) achieves much worse accuracy than its positive counterpart (M3). 
Including both positive and negative prompts (M4) performs better than a single prompt. 
When adding the evidential prompt, triple prompts (M5) achieve even higher accuracy, which indicates that \ours can better handle the complementary and beneficial information from the positive and negative sides. In Fig.~\ref{fig:ablation}(a)(b)(c), we analyze the impact of different lengths of prompt context in all three different experiment scenarios. In MLR with partial labels, we learn class-specific prompts and thus a smaller  $N$ (e.g. 12) in \ours  can perform well. For zero-shot learning in MLR, we learn uniform prompts shared by all classes, and it requires a larger N (e.g. 36) for good performance. In the main paper, we use N = 12 for all MLR experiments with partial labels and use N = 42 for experiments in zero-shot learning to keep a similar size to \ourconf.

\begin{table}[t]
    \begin{center}
     \caption{\small \textbf{Comparison between multi-headed attention and class-specific feature aggregation on MS-COCO} }~\label{table:mha_vs_conv_proj}

        \resizebox{\linewidth}{!}{
        \begin{tabular}{c| c c  | c c c c c  }
            \Xhline{3\arrayrulewidth} 
            Visual  &Fine-  & Train - Test  &   &  &  & &   \\
            Aggre. &tune. &  Resolution & $10\%$  & $30\%$ & $50\%$ & $70\%$ & $90\%$  \\
            \Xhline{3\arrayrulewidth} 
              \multirow{4}{*}{\makecell{Multi\\Headed \\ Atten.~\cite{vaswani2017attention}}}&  \xmark & 224 -- 224 & 70.4 & 74.1 & 74.8 & 75.4 & 75.7 \\
            &  \xmark  & 224 -- 448 & 65.9 & 70.2 & 71.2 & 72.0 & 72.1 \\
            &  \xmark  & 448 -- 448 & 72.1 & 75.5 & 76.5 & 77.1 & 77.3 \\
            &  \cmark  & 448 -- 448 & 74.1 & 77.6 & 78.2 & 78.5 & 78.4 \\
            \hline
              \multirow{4}{*}{\makecell{Evi.-Guided \\ Spatial\\Aggre.}} & \xmark  & 224 -- 224 &76.2	&78.7	&79.3	&79.7	&80.2 \\
            & \xmark  & 224 -- 448 &78.2     &79.5  &81.3  &82.1 &83.0\\
            & \xmark  & 448 -- 448 &81.4	 &83.7	&84.4  &84.8 &85.1\\
            & \cmark  & 448 -- 448 &81.9     &84.1  &85.2  &85.2 &85.7   \\
             \Xhline{3\arrayrulewidth} 
        \end{tabular}
        } 
    \end{center}
\end{table}

\vspace{1mm}
\noindent\textbf{Impact of Aggregation Function.}
In Table~\ref{table:mha_vs_conv_proj}, we compare the adaptive ability of different spatial aggregation methods. In general, training/testing with a larger resolution is beneficial as spatial details matter. 
As shown, multi-headed attention is bonded to the pre-training image resolution (224 in CLIP), while our evidence-guided region aggregation benefits from the increased input resolution either during training or in inference.  Our method uses original weights, but actually performs much better than finetuning the original multi-headed attention layer. 
We also compare different functions to aggregate the regional logits for each class in Fig.~\ref{fig:ablation} (d). We compute the final logits in four ways: (1) taking the average of logits at all spatial locations (``Ave''), (2) taking the region with the largest positive logit (``Max''), (3) generating aggregating weights for all spatial locations via a softmax function over the positive logits (``DualCoOp''), (4) generating aggregating weights for all spatial locations via a softmax function over the evidential logits (``DualCoOp++'').  ``Max'' performs better than ``Ave'', which indicates that the regional feature is more informative than the global feature in multi-label recognition. Taking into account both regional and global characteristics, ``DualCoOp'' improves performance. When further introducing the evidential, our ``DualCoOp++'' gives the best performance, which demonstrates that generating aggregating weights over the evidential logits  is better than aggregating via
positive logits. Visualization of different logits in  \ourconf and \ours is presented in Fig.~\ref{fig:result}.

\vspace{1mm}
\noindent\textbf{Ablation of Finetuning DualCoOp.} To better examine the effectiveness of finetuning for \ours, we also conduct experiments to finetune the weights in evidence-guided region aggregation. As shown in Table~\ref{table:mha_vs_conv_proj}, finetuning aggregation function can stably improve the performance over the non-finetuning setting with different amounts of labels available. We also tried finetuning all weights in the CLIP image encoder, yet find that performance drops significantly especially when given a lower portion of training labels, which shows that tuning with insufficient supervision may undermine the pretrained vision-language alignment in CLIP models.


\section{Conclusion}

In this paper,  we extend our previous \ourconf with a novel framework \ours, unified for two types of multi-label recognition with limited annotations: partial-label and zero-shot. \ours utilizes a powerful vision-language pretraining model obtained from a web-scale dataset. 
By introducing a lightweight learnable overhead, it can quickly adapt to solve multi-label recognition after receiving a small amount of labels. 
In \ours, we learn a triplet of evidential, positive, and negative prompts followed by the target class name as linguistic input. Furthermore, to better aggregate visual region features for each class, we reformulate the original visual attention in the pretraining model as an evidence-guided region feature aggregation. Moreover, a winner-take-all module is introduced to promote the cross-label interaction and regularize that each region positively responds to at most one class. We conduct extensive experiments for both partial-label MLR and Zero-Shot MLR across MS-COCO, VOC2007, and NUS-WIDE datasets, showing the improvements of \ours to \ourconf and the efficacy of our proposed approach over state-of-the-art methods.


\bibliographystyle{IEEEtran}
\bibliography{egbib}

\end{document}